\documentclass[letterpaper, 10 pt, conference]{ieeeconf}  

\IEEEoverridecommandlockouts                              

\usepackage{bm}
\usepackage{url}
\usepackage{graphicx}
\usepackage{lipsum}
\usepackage{amsmath}
\usepackage{amssymb}
\usepackage{nccmath}
\usepackage{comment}
\usepackage{balance}
\usepackage{multirow}
\usepackage{textcomp}
\usepackage{multirow}
\usepackage{subcaption}
\usepackage{booktabs}
\usepackage{algorithm}
\usepackage{dblfloatfix}
\usepackage[table]{xcolor}
\usepackage[noend]{algpseudocode}

\title{\LARGE \bf
Globally Consistent and Tightly Coupled 3D LiDAR Inertial Mapping
}

\author{Kenji Koide$^{1}$, Masashi Yokozuka$^{1}$, Shuji Oishi$^{1}$, and Atsuhiko Banno$^{1}$
\thanks{This work was supported in part by a project commissioned by the New Energy and Industrial Technology Development Organization (NEDO).}
\thanks{$^{1}$All the authors are with the Department of Information Technology and Human Factors, the National Institute of Advanced Industrial Science and Technology, Umezono 1-1-1, Tsukuba, 3050061, Ibaraki, Japan, {\tt\small k.koide@aist.go.jp}}%
}

\begin{document}

\maketitle
\thispagestyle{empty}
\pagestyle{empty}

\setlength\floatsep{8pt}
\setlength\textfloatsep{8pt}

\newcommand\blfootnote[1]{%
  \begingroup
  \renewcommand\thefootnote{}\footnote{#1}%
  \addtocounter{footnote}{-1}%
  \endgroup
}

\begin{abstract}

This paper presents a real-time 3D mapping framework based on global matching cost minimization and LiDAR-IMU tight coupling. The proposed framework comprises a preprocessing module and three estimation modules: odometry estimation, local mapping, and global mapping, which are all based on the tight coupling of the GPU-accelerated voxelized GICP matching cost factor and the IMU preintegration factor. The odometry estimation module employs a keyframe-based fixed-lag smoothing approach for efficient and low-drift trajectory estimation, with a bounded computation cost. The global mapping module constructs a factor graph that minimizes the global registration error over the entire map with the support of IMU constraints, ensuring robust optimization in feature-less environments. The evaluation results on the Newer College dataset and KAIST urban dataset show that the proposed framework enables accurate and robust localization and mapping in challenging environments.

\end{abstract}

\section{Introduction}

Environmental mapping is an inevitable function of autonomous systems and LiDAR is one of the most common sensors used for mapping tasks owing to its ranging accuracy and reliability. Following recent visual SLAM studies, tightly coupled LiDAR-IMU fusion techniques have been widely studied in recent years \cite{Ye2019,Qin2020}. The tight coupling scheme fuses LiDAR and IMU measurements on a unified objective function and makes the sensor trajectory estimation robust to quick sensor motion, as well as feature-less environments, where sufficient geometrical constraints are not available. Furthermore, IMU measurements provide information on the direction of gravity, which enables a reduction of the estimation drift \cite{Qin2018}.


However, the use of the LiDAR-IMU tight coupling scheme has mostly been limited to the frontend (i.e., odometry estimation) of the system in the context of LiDAR SLAM. This is because the backend (i.e., global optimization) of most existing methods relies on pose graph optimization; this uses approximated relative pose constraints constructed from the estimation result of the frontend, resulting in the separation of LiDAR- and IMU-based estimation.

In this paper, we propose a real-time SLAM framework that employs a tightly coupled LiDAR-IMU fusion scheme for all estimation stages (i.e., from odometry estimation to global optimization). We use the voxel-based GICP matching cost factor, which can fully leverage GPU parallel processing and enables the creation of a factor graph to minimize the scan matching error over the entire map \cite{koide_ral2021}. We combine the GPU-accelerated matching cost factor with the IMU preintegration factor to jointly consider the LiDAR and IMU constraints for global trajectory optimization. This approach enables us to accurately correct estimation drift in challenging environments while preserving the global consistency of the map. To the best of our knowledge, this is the first study to perform global trajectory optimization based on the tight coupling of LiDAR and IMU constraints. We also propose a keyframe-based LiDAR-IMU frontend algorithm with fixed-lag smoothing that enables efficient and low-drift sensor ego-motion estimation with a bounded computation cost. We show that the proposed framework enables highly accurate and robust trajectory estimation through experiments on the Newer College dataset \cite{Ramezani2020} and KAIST urban dataset \cite{Jeong2018}. 

The proposed framework is distinct from existing LiDAR-IMU SLAM frameworks in several aspects.
\begin{enumerate}
  \item It is based on the voxelized GICP matching cost factor \cite{koide_ral2021}, which uses a larger number of points to calculate the registration error compared to the commonly used scan matching based on line and plane point matching \cite{Ye2019}. This enables accurate and robust constraint of sensor poses while fully leveraging GPU parallel processing.
  \item Its tightly coupled odometry estimation module employs a keyframe-based fixed-lag smoothing method inspired by \cite{Engel2018}, which enables a low-drift trajectory estimation with a bounded computation cost.
  \item It also employs the tight coupling approach for the backend. The backend constructs a densely connected matching cost factor graph with the support of the IMU factors and exhibits outstanding accuracy. It also introduces the concept of {\it endpoints} of submaps to strongly constrain submaps at a large time interval with IMU constraints.
\end{enumerate}

\section{Related Work}

\subsection{LiDAR-IMU frontend}

Following the recent progress in visual-inertial SLAM techniques \cite{Qin2018,Stumberg2018,Campos2021}, LiDAR-IMU fusion has been an important topic for LiDAR SLAM \cite{liosam2020shan}. The use of IMU enables us to predict sensor motion at a frequency of 100-1000 Hz, facilitating good initial estimates of the sensor pose and correct distortion of LiDAR points under quick sensor motion. Furthermore, IMU measurements provide information on the direction of gravity, enabling a reduction of the trajectory estimation drift in four DoFs by aligning the trajectory with this direction \cite{Qin2018}.

One method for fusing IMU and LiDAR measurements is the loose coupling scheme, which separately considers LiDAR-based estimation and IMU-based estimation and fuses the estimation results in the pose space using, for example, an extended Kalman filter \cite{Weiss2011} or a factor graph \cite{liosam2020shan,Indelman2013}. While the loose coupling scheme is computationally efficient, another approach, i.e., the tight coupling scheme, can theoretically be more accurate and robust than the loose coupling scheme \cite{Ye2019}. 

The tight coupling scheme fuses LiDAR and IMU measurements on a unified objective function. This approach enables robust estimation of sensor trajectory in feature-less environments, where sufficient geometrical information is not available through LiDAR data, because IMU constraints help to constrain the sensor trajectory based on inertial information. Owing to their high accuracy and robustness, tightly coupled LiDAR-IMU methods have been widely studied in recent years \cite{Ye2019,Qin2020,Xu2021,Li2021}. 

Despite its theoretical advantages, the tight coupling approach considerably increases system complexity and computational cost and can be unstable in extreme situations. To avoid increasing system complexity, several methods employ an IMU-centric loose coupling approach to make the system robust in extreme environments (e.g., in an underground environment) \cite{Palieri2021,zhao2021super}.

\subsection{LiDAR-IMU backend}

While there are many LiDAR SLAM frontend methods based on LiDAR-IMU fusion, the use of IMU constraints is mostly limited to the frontend (i.e., odometry estimation) in most existing methods \cite{liosam2020shan,Li2021,Shan2018} because they use pose graph optimization for global trajectory optimization, which minimizes errors in the pose space. As pose graph optimization uses SE3 relative pose constraints to constrain sensor poses, separation of the LiDAR-based estimation and IMU-based estimation is unavoidable. It also affects the consistency of the map when closing a large loop or constraining small overlapping frames because it employs an approximated representation (i.e., Gaussian distribution) for relative pose constraints \cite{koide_ral2021}.

The backend of the proposed framework is conceptually similar to Voxgraph \cite{Reijgwart2020}, which also considers point cloud registration errors for global trajectory optimization. It uses the Euclidean signed distance field \cite{Oleynikova2017} to represent submaps and efficiently computes the registration error between submaps, without requiring a costly nearest-neighbor search. However, the registration error minimization was still computationally expensive, and global optimization was conducted using a random subset of registration residuals with the support of SE3 relative pose factors. The proposed method eliminates inaccurate SE3 relative pose factors and fully relies on matching cost factors for all optimization stages, resulting in globally consistent mapping results. Furthermore, it also enables the construction of a tightly coupled global trajectory optimization, which greatly improves the robustness of the mapping process in severely feature-less environments.

\section{Methodology}

\begin{figure}[tb]
  \centering
  \includegraphics[width=0.8\linewidth]{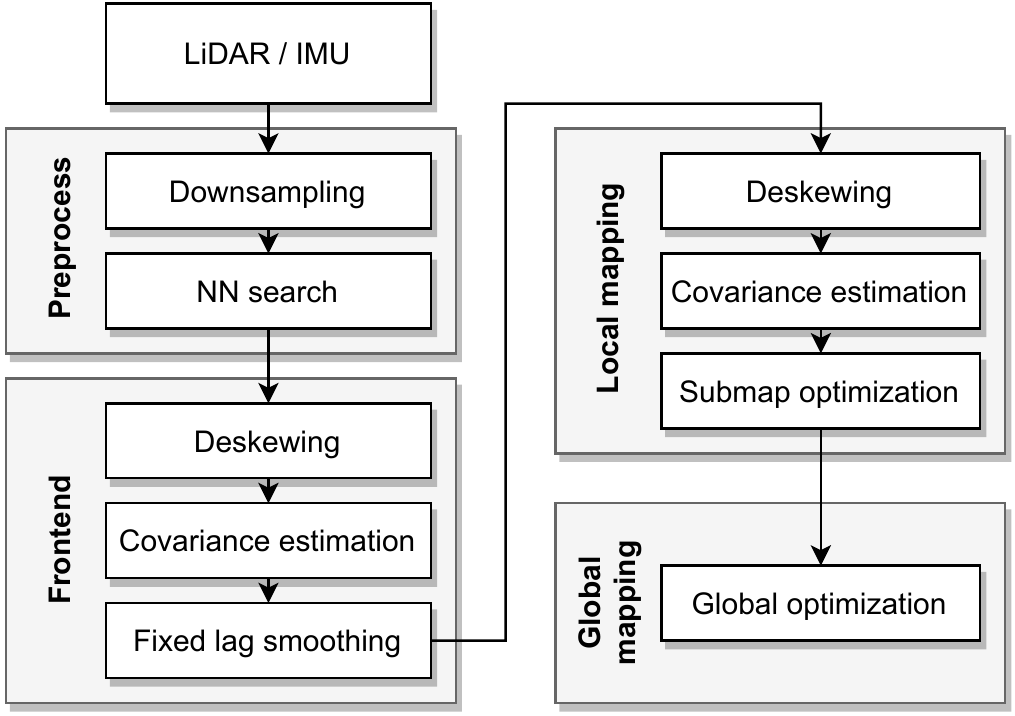}
  \caption{System overview.}
  \label{fig:system}
\end{figure}

Fig. \ref{fig:system} shows an overview of the proposed framework, comprising a preprocessing module and three estimation modules, i.e., odometry estimation, local mapping, and global mapping, which are all based on tightly coupled LiDAR-IMU fusion. The odometry estimation (i.e., frontend) module robustly estimates the sensor motion and provides an initial estimate of the latest sensor state. The estimated sensor states are refined by the following local mapping module, and several local frames are merged into one submap. The global mapping module then optimizes the submap poses such that the global registration error is minimized while preserving the consistency of the map. We run these modules in parallel via multi-threading.

We define the sensor state ${\bm x}_t$ that will be estimated in the estimation modules as

\begin{align}
{\bm x}_t = [{\bm T}_t, {\bm v}_t, {\bm b}_t]^T,
\end{align}
where ${\bm T}_t = [{\bm R}_t | {\bm t}_t] \in SE(3)$ is the sensor pose, ${\bm v}_t \in \mathbb{R}^3$ is the velocity, and ${\bm b}_t = [{\bm b}_t^a, {\bm b}_t^{\omega}] \in \mathbb{R}^6$ is the IMU acceleration and angular velocity bias. We estimate the time series of sensor states from LiDAR point clouds $\mathcal{P}_t$ and IMU measurements (linear acceleration $a_t$ and angular velocity $\omega_t$). Note that we transform LiDAR point clouds into the IMU coordinate frame and, for efficiency and simplicity, consider them as if they are in a unified sensor coordinate frame.

In the following Sec. \ref{sec:matching_cost_factor} and \ref{sec:imu_preintegration_factor}, we first introduce two types of factors, the LiDAR matching cost factor and the IMU preintegration factor, that are the main components of the factor graphs used in the proposed framework. Then, we explain each module in the proposed framework in Sec. \ref{sec:preprocess} to \ref{sec:global_mapping}.

\subsection{LiDAR Matching Cost Factor}
\label{sec:matching_cost_factor}

The matching cost factor constrains two sensor poses (${\bm T}_i$ and ${\bm T}_j$) such that the matching cost between the point clouds ($\mathcal{P}_i$ and $\mathcal{P}_j$) is minimized. As the matching cost, we choose the voxelized GICP (VGICP) cost \cite{vgicp}, which is a variant of generalized ICP \cite{Segal2009} suitable for GPU computation. 

VGICP models each input point ${\bm p}_k \in \mathcal{P}_i$ as a Gaussian distribution ${\bm p}_k = ({\bm \mu}_k, {\bm C}_k$), and the covariance matrix ${\bm C}_k$ is computed from the neighboring points of ${\bm p}_k$. It discretizes $\mathcal{P}_j$ into voxels and computes a Gaussian distribution for each voxel by aggregating the means and covariances of the points in the voxel. Then, the matching cost $e^M$ between $\mathcal{P}_i$ and $\mathcal{P}_j$ is defined based on the GICP distribution-to-distribution distance:

\begin{align}
\label{eq:vgicp}
e^M(\mathcal{P}_i, \mathcal{P}_j, {\bm T}_i, {\bm T}_j) &= \sum_{p_k \in \mathcal{P}_i} e^{\text{\it D2D}}({\bm p}_k, {\bm T}_i^{-1} {\bm T}_j), \\
e^{\text{\it D2D}} ({\bm p}_k, {\bm T}_{ij}) &= {\bm d}_k^T ({\bm C}'_k + {\bm T}_{ij}{\bm C}_k{\bm T}_{ij}^T)^{-1} {\bm d}_k,
\end{align}
where ${\bm p}_k' = ({\bm \mu}_k', {\bm C}_k')$ is the mean and covariance of the corresponding voxel of ${\bm p}_k$ given by looking up the voxel map of $\mathcal{P}_j$, and ${\bm d}_k = {\bm \mu}_k' - {\bm T}_{ij} {\bm \mu}_k$ is the residual between ${\bm \mu}_k$ and ${\bm \mu}'_k$.

From the derivatives of Eq. \ref{eq:vgicp}, we obtain a Hessian factor to constrain the relative pose between ${\bm T}_i$ and ${\bm T}_j$. It is worth emphasizing that we re-evaluate and linearize $e^M$ at the current linearization point for every optimization iteration, which results in a more accurate constraint than the traditional SE3 relative pose constraint \cite{koide_ral2021}.

\subsection{IMU Preintegration Factor}
\label{sec:imu_preintegration_factor}

We use the IMU preintegration technique \cite{Forster2017} to efficiently incorporate IMU constraints into the factor graph. Given an IMU measurement (${\bm a}_t$ and ${\bm \omega}_t$), the sensor state evolves as follows:

\begin{align}
\label{eq:imu_evol_R}
{\bm R}_{t + \Delta t} &= {\bm R}_t \exp \left( \left( {\bm \omega}_t - {\bm b}_t^{\omega} - {\bm \eta}_k^{\omega} \right) \Delta t \right), \\
\label{eq:imu_evol_v}
{\bm v}_{t + \Delta t} &= {\bm v}_t + {\bm g} \Delta t + {\bm R}_t \left( {\bm a}_t - {\bm b}_t^a - {\bm \eta}_t^a \right) \Delta t, \\
\label{eq:imu_evol_p}
{\bm t}_{t + \Delta t} &= {\bm t}_t + {\bm v}_t \Delta t + \frac{1}{2} {\bm g} \Delta t^2 + \frac{1}{2} {\bm R}_t \left( {\bm a}_t - {\bm b}_t^a - {\bm \eta}_t^a \right) \Delta t^2,
\end{align}
where ${\bm g}$ is the gravity vector and ${\bm \eta}_t^a$ and ${\bm \eta}_t^{\omega}$ are white noise in the IMU measurement.

The IMU preintegration factor integrates the system evolution between two time steps $i$ and $j$ to obtain the relative body motion constraints (see \cite{Forster2017} for a detailed derivation):

\begin{align}
\label{eq:preint_R}
\Delta {\bm R}_{ij} &= {\bm R}_i^T {\bm R}_j \exp \left( \delta {\bm \phi}_{ij} \right), \\
\Delta {\bm v}_{ij} &= {\bm R}_i^T \left( {\bm v}_j - {\bm v}_i -{\bm g} \Delta t_{ij} \right) + \delta {\bm v}_{ij}, \\
\Delta {\bm t}_{ij} &= {\bm R}_i^T \left( {\bm t}_j - {\bm t}_i - {\bm v} \Delta t_{ij} - \frac{1}{2} {\bm g} \Delta t_{ij}^2 \right) + \delta {\bm t}_{ij},
\end{align}
where $\delta {\bm \phi}_{ij}, \delta {\bm v}_{ij}$, and $\delta {\bm p}_{ij}$ are white noise in the integration process. 

The IMU preintegration factor enables us to keep the factor graph well-constrained in environments where geometrical features are insufficient and LiDAR factors can be deficient. Furthermore, it provides information on the direction of gravity and reduces the estimation drift in 4 DoF \cite{Qin2018}.

\subsection{Preprocessing}
\label{sec:preprocess}

We first downsample the input point clouds with a voxel grid filter. For the following deskewing process, we average the timestamps of points in addition to the positions for each voxel. 
If a point has a timestamp that is significantly different from that of the corresponding voxel (e.g., $|t^{\text{\it point}} - t^{\text{\it voxel}}| > \frac{d^{\text{\it scan}}}{10}$, where $d^{\text{\it scan}}$ is the scan duration), we assign the point to another new voxel to avoid fusing the first and last points of a scan. We then find k neighboring points for each point required for the subsequent point covariance estimation. We assume that the neighborhood relationship of points does not largely change during the following deskewing process and use the precomputed nearest neighbors in covariance estimation, which is performed after deskewing.

\subsection{Odometry Estimation}
\label{sec:odometry}

\begin{figure}[tb]
  \centering
  \includegraphics[width=1.0\linewidth]{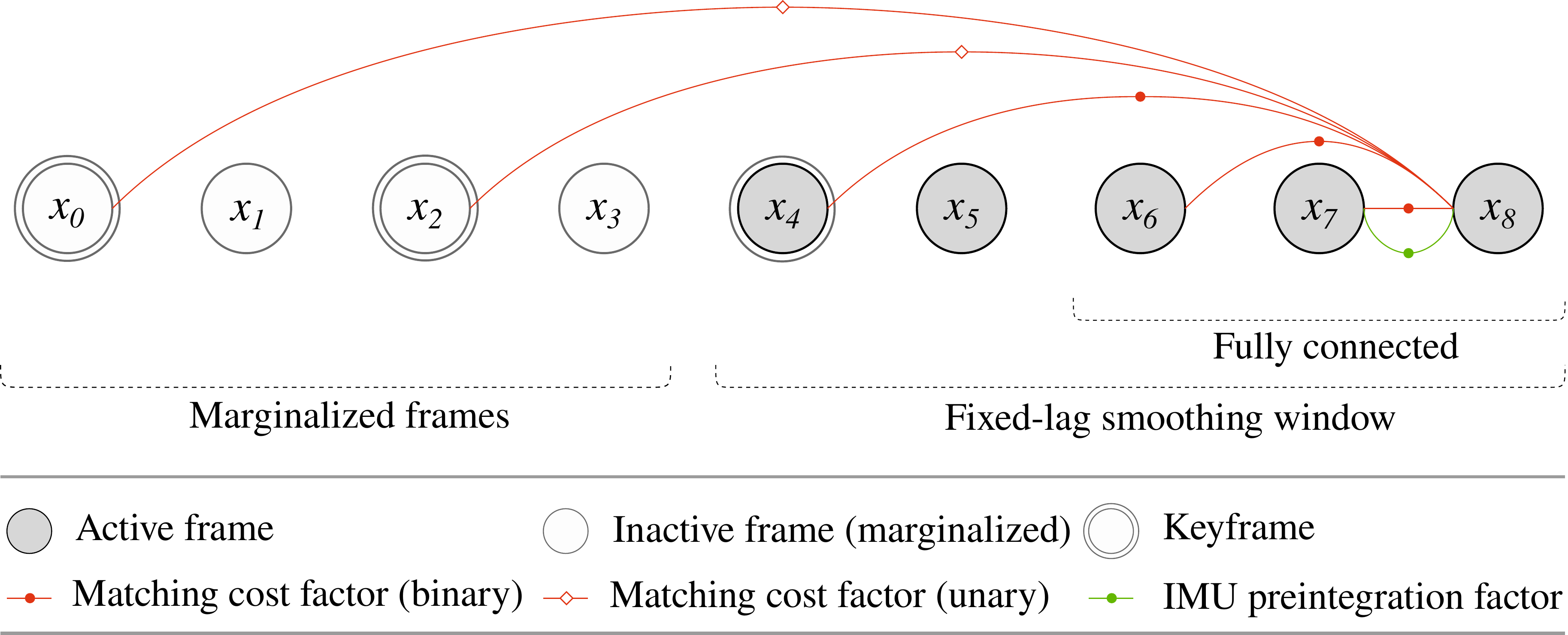}
  \caption{Frontend factor graph. Only the factors related to the latest frame ($x_8$) are illustrated. Matching cost factors are created for the last $N$ frames and keyframes. If a keyframe is outside the fixed-lag smoothing window (is already marginalized out), we create a unary matching cost factor. IMU preintegration factors are created between consecutive frames.}
  \label{fig:frontend_graph}
\end{figure}

The odometry estimation module compensates for quick sensor motion and robustly estimates the sensor state by fusing LiDAR and IMU measurements. We first correct the distortion on the point cloud caused by the sensor motion by transforming the points into the IMU frame with motion prediction based on IMU dynamics. We then compute the covariance of each point using the precomputed neighboring points.

Given the deskewed point clouds, we construct the factor graph shown in Fig. \ref{fig:frontend_graph}. To limit computation cost and ensure that the system is real-time capable, we use a fixed-lag smoothing approach and marginalize the old frames. Inspired by direct sparse odometry \cite{Engel2018}, we introduce a keyframe mechanism for efficient and low-drift trajectory estimation. Keyframes are a set of frames that are selected such that they are spatially well-distributed while having sufficient overlap with the latest frame. We create a matching cost factor between the latest frame and every keyframe to efficiently reduce estimation drift. If a keyframe is already marginalized from the fixed-lag smoother, we consider the keyframe pose as fixed and create a unary matching cost factor that constrains the latest sensor pose with respect to the fixed keyframe.

To manage keyframes, we define an overlap rate between two frames $\mathcal{P}_i$ and $\mathcal{P}_j$ as the fraction of points in $\mathcal{P}_i$ that fall within a voxel of $\mathcal{P}_j$ \cite{koide_ral2021}. Every time a new frame arrives, we evaluate the overlap rate between that frame and the latest keyframe and, if the overlap is smaller than a threshold (e.g., 90\%), we insert that frame into the keyframe list. Similar to the keyframe marginalization strategy in \cite{Engel2018}, we remove redundant keyframes using the following strategy:

\begin{enumerate}
  \item We remove keyframes that overlap the latest keyframe by less than a certain threshold (e.g., 5\%).
  \item If more than $N^{\text{\it odom}}$ (e.g., 20) frames exist in the keyframe list, we remove the keyframe that minimizes the following score:
    \begin{align}
      s(i) = o(i, N^{\text{\it odom}}) \sum_{j \in [1, N^{\text{\it odom}}-1] \backslash \{i\}} \left( 1 - o(i, j) \right),
    \end{align}
    where $o(i, j)$ is the overlap rate between the i-th and j-th keyframes. The score function is heuristically designed to keep keyframes spatially well-distributed while leaving more keyframes close to the latest one.
\end{enumerate}

In addition to the keyframes, we create matching cost factors between the latest frame and the last few frames (e.g., last three frames) to make the odometry estimation robust to quick sensor motion. We also create an IMU preintegration factor between consecutive frames for robustness in feature-less environments. 

\subsection{Local Mapping}
\label{sec:local_mapping}

\begin{figure}[tb]
  \centering
  \includegraphics[width=1.0\linewidth]{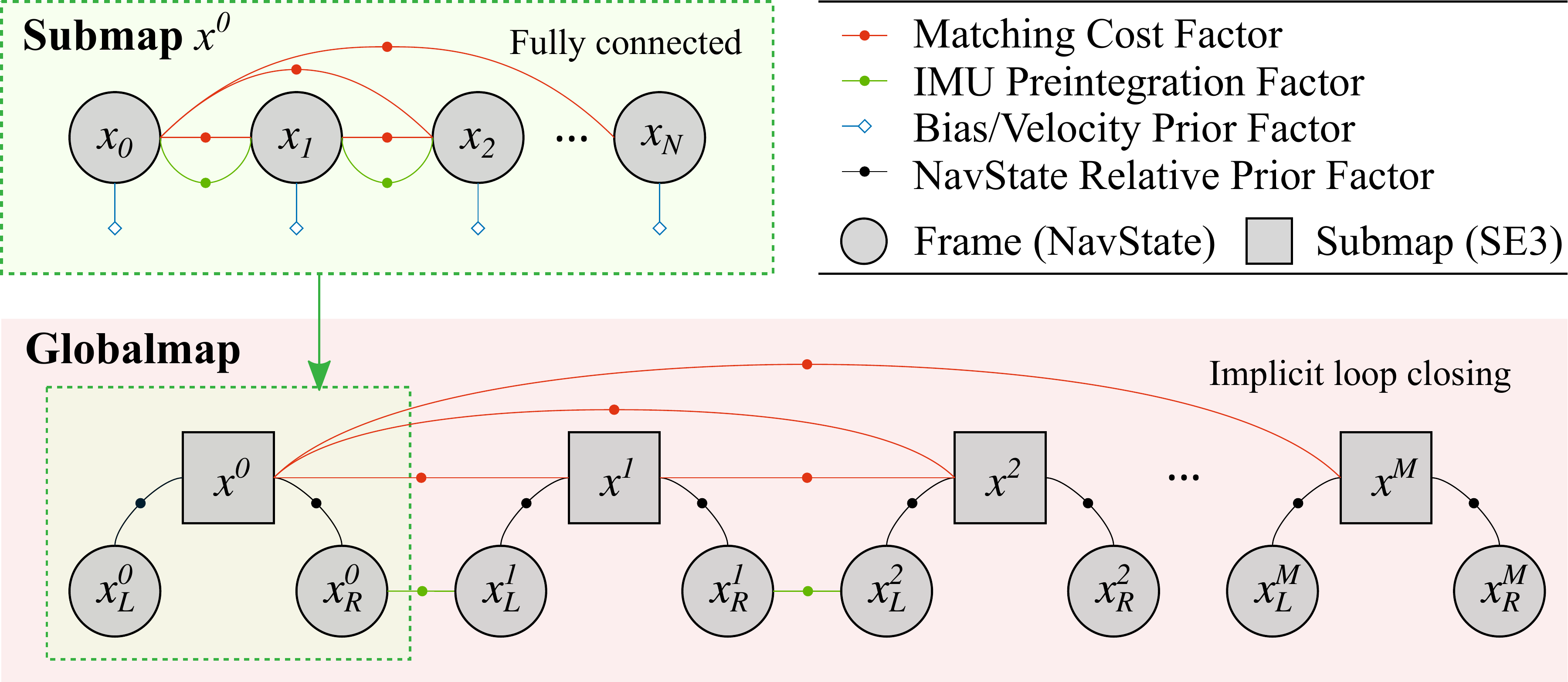}
  \caption{Backend factor graph. The local mapping module merges several local frames into one submap using an all-to-all registration strategy. The global mapping module optimizes the submap poses such that the global registration error is minimized over the entire map. Both modules take advantage of IMU factors to stabilize the estimation in severe feature-less environments and reduce estimation drift.}
  \label{fig:backend_graph}
\end{figure}

Once a frame is marginalized from the odometry estimation graph, it is fed to the local mapping module as an initial estimate of the sensor state. The local mapping module merges several local frames into one submap to reduce the number of optimized variables in the global mapping module.

We first re-perform deskewing and covariance estimation with the marginalized state, which is expected to improve upon the initial prediction made at the beginning of the odometry estimation. We then evaluate the overlap rate between that frame with the latest frame in the submap and, if the overlap rate is smaller than a threshold (e.g., 90\%), insert that frame into the submap factor graph.

As shown in Fig. \ref{fig:backend_graph}, we create a matching cost factor for every combination of frames in the submap (i.e., all-to-all registration). We also add an IMU preintegration factor between consecutive frames and add a prior factor for the velocity and bias of each frame, based on the marginalized state, to better stabilize the submap optimization.

Once the number of frames in the submap becomes equal to $N^{\text{\it sub}}$ (e.g., 15) or the overlap between the first and last frames becomes smaller than a threshold (e.g., 5 \%), we perform factor graph optimization using the Levenberg-Marquardt optimizer \cite{Levenberg1944} and merge the frames into one submap based on the optimization result.

\subsection{Global Mapping}
\label{sec:global_mapping}

The global mapping module corrects the estimation drift to obtain a globally consistent mapping result. We create a matching cost factor between every submap pair with an overlap rate exceeding a small threshold (e.g., 5\%). This results in an extremely dense factor graph. Every submap is aligned with not only adjacent submaps on the graph but also every revisited submap that results in closing loops implicitly.

Submaps are created at a larger time interval (e.g., 10 s). If we simply create an IMU factor between submaps, its uncertainty becomes too large, and it cannot strongly constrain the relative pose between submaps \cite{Stumberg2018,MurArtal2017a}. Furthermore, we also lose information on the velocity and IMU bias estimated by the local mapping module. To address these problems, we introduce two states called {\it endpoints} (${\bm x}^i_L$ and ${\bm x}^i_R$) for each submap ${\bm x}^i$; they hold the states of the first and last frames in the submap with respect to the submap pose.

Given an estimate of sensor states $[{\bm x}_0, \cdots, {\bm x}_{N^{\text{\it sub}}}]$ in a submap ${\bm x}^i$, we define the submap origin ${\bm T}^i$ as the pose of the sensor pose at the center ${\bm T}_{N^{\text{\it sub}}/2}$. Then, the sensor state ${\bm x}_t$ relative to the submap origin is given as:

\begin{align}
\label{eq:relative_T}
{\bm T}'_t &= \left( {\bm T}^i \right)^{-1} {\bm T}_t, \\
\label{eq:relative_v}
{\bm v}'_t &= \left( {\bm R}^i \right)^{-1} {\bm v}_t, \\
\label{eq:relative_b}
{\bm b}'_t &= {\bm b}_t.
\end{align}

We create relative state factors between a submap ${\bm x}^i$ and endpoints ${\bm x}^i_L$ and ${\bm x}^i_R$ respectively from the first and last frames in the submap (${\bm x}_0$ and ${\bm x}_{N^{\text{\it sub}}}$) such that they satisfy the relative state relationship described by Eqs. \ref{eq:relative_T} - \ref{eq:relative_b}. We then create an IMU factor between ${\bm x}^i_R$ and ${\bm x}^{i+1}_L$. In this way, an IMU factor covers a small time interval and can strongly constrain the submap poses while avoiding the loss of the velocity and bias information estimated by the local mapping module. 

Every few times a new submap is inserted (e.g., every five submaps), the factor graph is incrementally optimized via the iSAM2 optimizer \cite{Kaess2011} in GTSAM\footnote{\url{https://gtsam.org/}}.

\section{Evaluation}

\subsection{Evaluation on the Newer College Dataset}

\begin{table*}[tb]
  \centering
  \caption{Evaluation results on the Newer College dataset}
  \label{tab:result_newer}
  \begin{tabular}{c|cccccc}
  \toprule
  Metric     & LIOM (odom)       & LIO-SAM (odom)    & LIO-SAM           & Proposed (odom)   & Proposed \\
  \midrule
  RTE [m]    & 2.224 $\pm$ 1.402 & 2.215 $\pm$ 1.376 & 2.156 $\pm$ 1.357 & {\bf 2.140} $\pm$ 1.348 & 2.160 $\pm$ 1.356  \\
  ATE [m]    & 3.392 $\pm$ 1.653 & 1.176 $\pm$ 0.641 & 0.529 $\pm$ 0.259 & 0.899 $\pm$ 0.595       & {\bf 0.276} $\pm$ 0.093  \\
  \bottomrule
  \end{tabular}
\end{table*}

\begin{figure}[tb]
  \centering
  \includegraphics[width=0.9\linewidth]{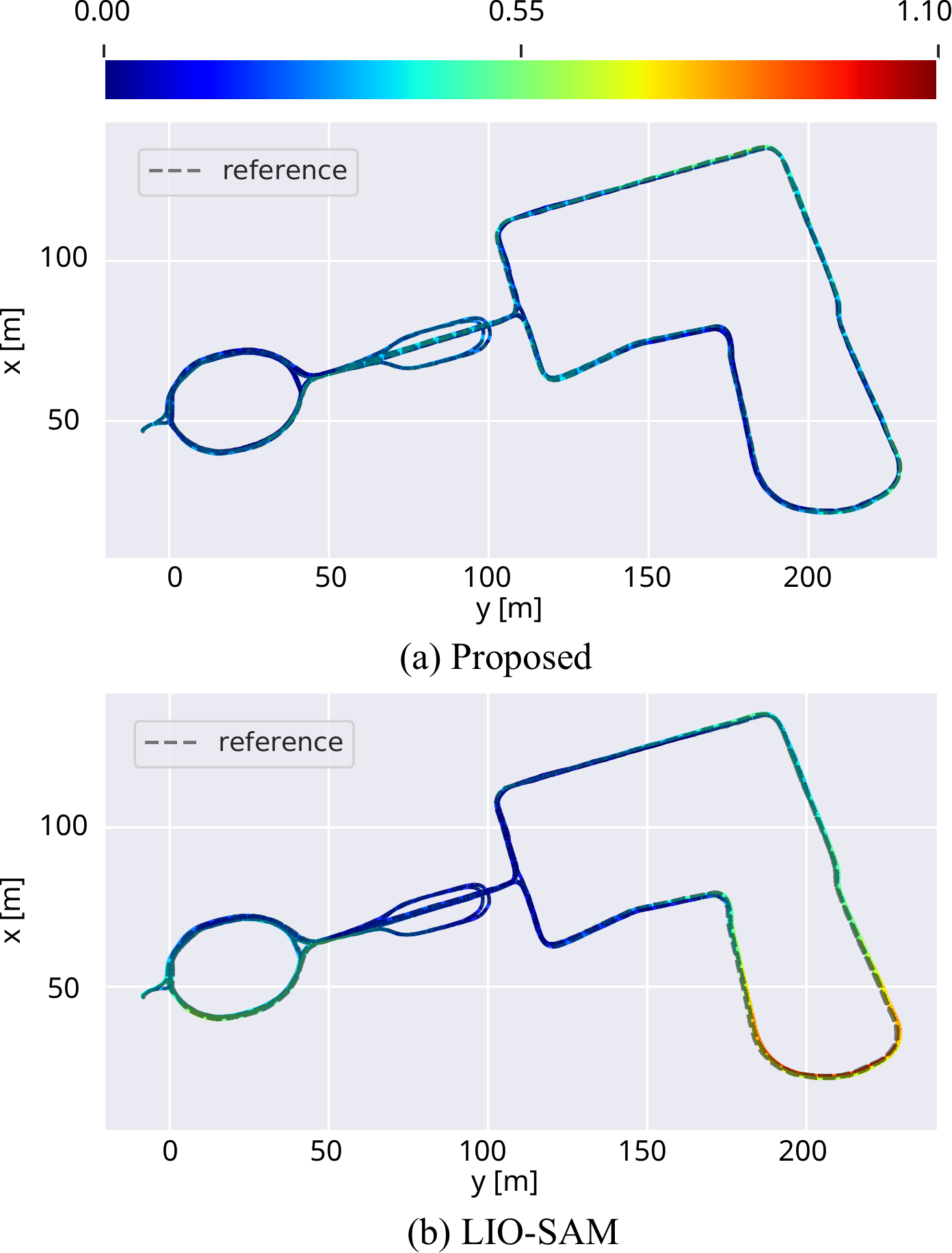}
  \caption{Sensor trajectories estimated by the proposed framework and LIO-SAM for the {\it long\_experiment} sequence in the Newer College dataset. The color indicates the magnitude of the ATE.}
  \label{fig:newer_trajs}
\end{figure}

We conducted experiments on the Newer College dataset \cite{Ramezani2020} recorded with an Ouster OS-1 64, which provides LiDAR point clouds at 10 Hz accompanied by synchronized IMU data at 100 Hz. We compared the proposed framework with two state-of-the-art LiDAR-IMU SLAM frameworks --- i.e., LIO-mapping (LIOM) \cite{Ye2019} and LIO-SAM \cite{liosam2020shan} --- on the {\it long\_experiment} sequence, which is the longest sequence in the Newer College dataset (3,060 m / 2,650 s). As the evaluation metric, we used the absolute trajectory error (ATE) and 100 m relative trajectory error (RTE) \cite{Zhang2018}. We used an Intel Core i9-9900K and Nvidia RTX 2080 for all the experiments.

Table \ref{tab:result_newer} presents the quantitative evaluation results. The proposed frontend algorithm showed a comparable RTE (2.140 m) to those of LIO-mapping and LIO-SAM (2.224 m and 2.215 m, respectively). This result suggests that the proposed keyframe-based odometry estimation enables a low-drift trajectory estimation.  With global optimization, the proposed framework greatly improved the trajectory consistency and demonstrated a significantly better ATE (0.276 m) than that of LIO-SAM (0.529 m).

Fig. \ref{fig:newer_trajs} shows the trajectories estimated by the proposed framework and LIO-SAM. LIO-SAM exhibited large errors on a large curve. We infer that is because 1) the density of the frames is relatively small at the corner and LIO-SAM failed to create sufficient relative pose constraints. 2) LIO-SAM does not incorporate IMU factors in the global optimization that resulted in losing the gravity direction information. Meanwhile, the proposed framework showed an accurate and consistent trajectory estimation result owing to the global matching cost minimization scheme and the tight coupling of LiDAR and IMU constraints. 

It is worth mentioning that the proposed method is very robust to quick sensor motion. We confirmed that it successfully estimated the sensor trajectory in {\it quad\_with\_dynamics} and {\it dynamic\_spinning} sequences, which presented aggressive sensor motion (up to 1.5 m/s and 3.5 rad/s) \footnote{See the supplementary video.}.

\subsection{Evaluation on KAIST Urban Dataset}

\begin{figure}[tb]
  \centering
  \includegraphics[width=0.35\linewidth]{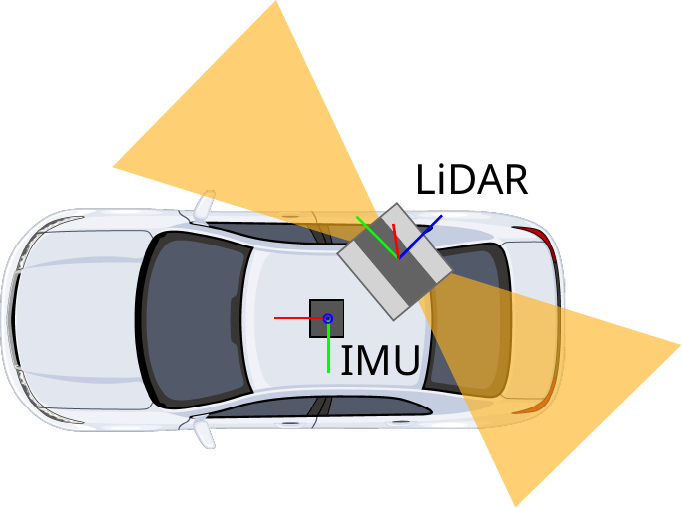}
  \caption{Sensor configuration of KAIST urban dataset.}
  \label{fig:kaist_urban}
\end{figure}

\begin{figure*}[tb]
  \centering
  \begin{minipage}[b]{0.44\linewidth}
  \centering
  \includegraphics[width=\linewidth]{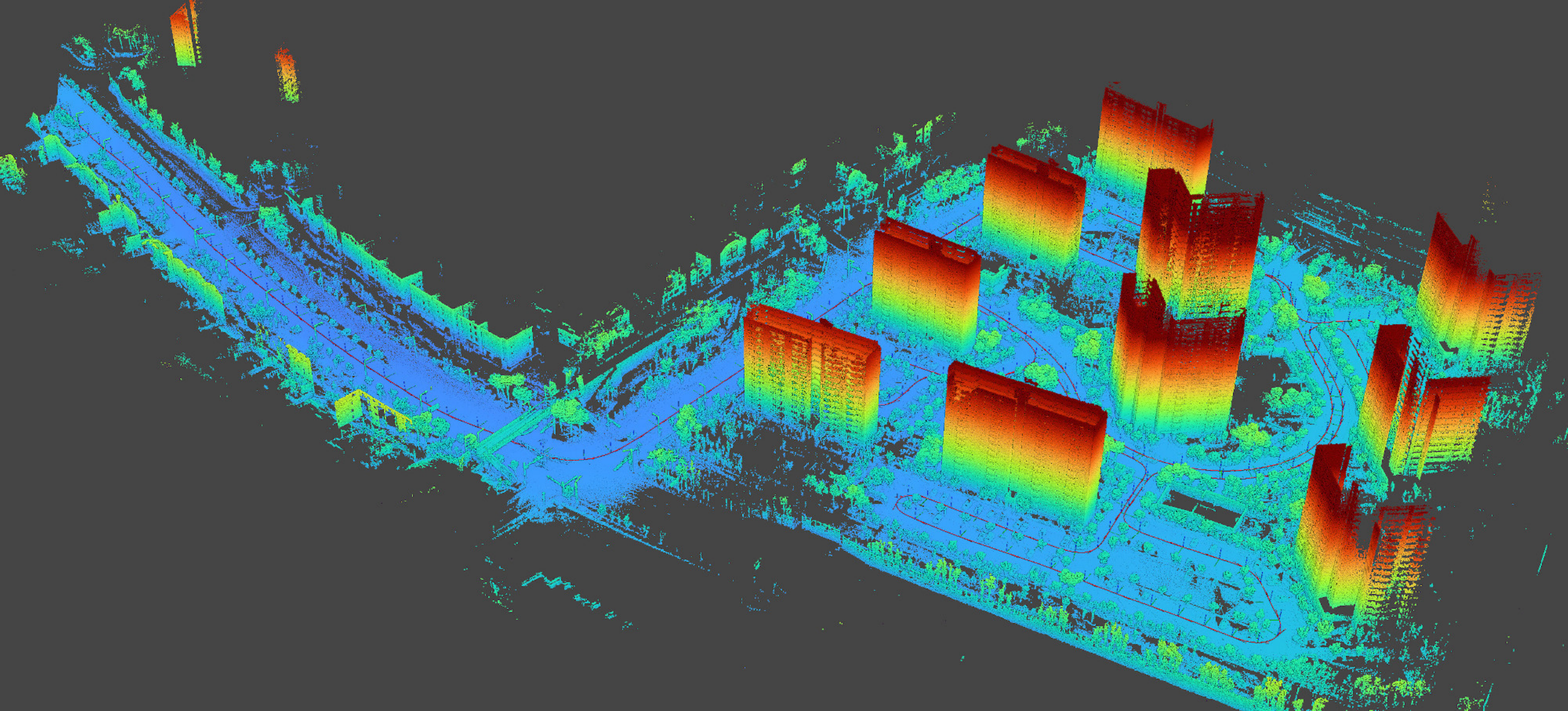}
  \subcaption{Estimated map and trajectory}
  \end{minipage}
  \begin{minipage}[b]{0.44\linewidth}
  \centering
  \includegraphics[width=\linewidth]{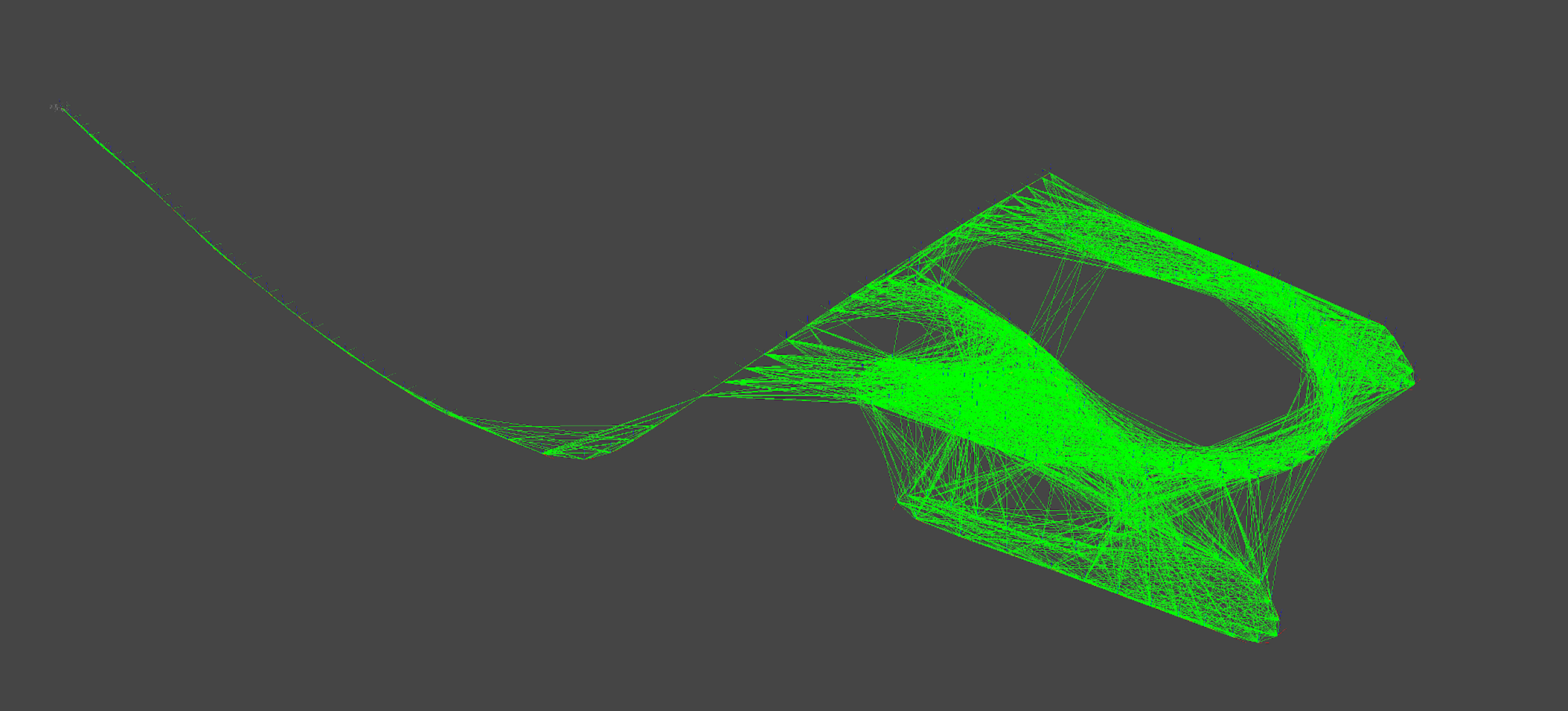}
  \subcaption{Factor graph}
  \end{minipage}
  \caption{Mapping result for the KAIST07 sequence.}
  \label{fig:kaist_07}
\end{figure*}

\begin{figure*}[tb]
  \centering
  \begin{minipage}[b]{0.44\linewidth}
  \centering
  \includegraphics[width=\linewidth]{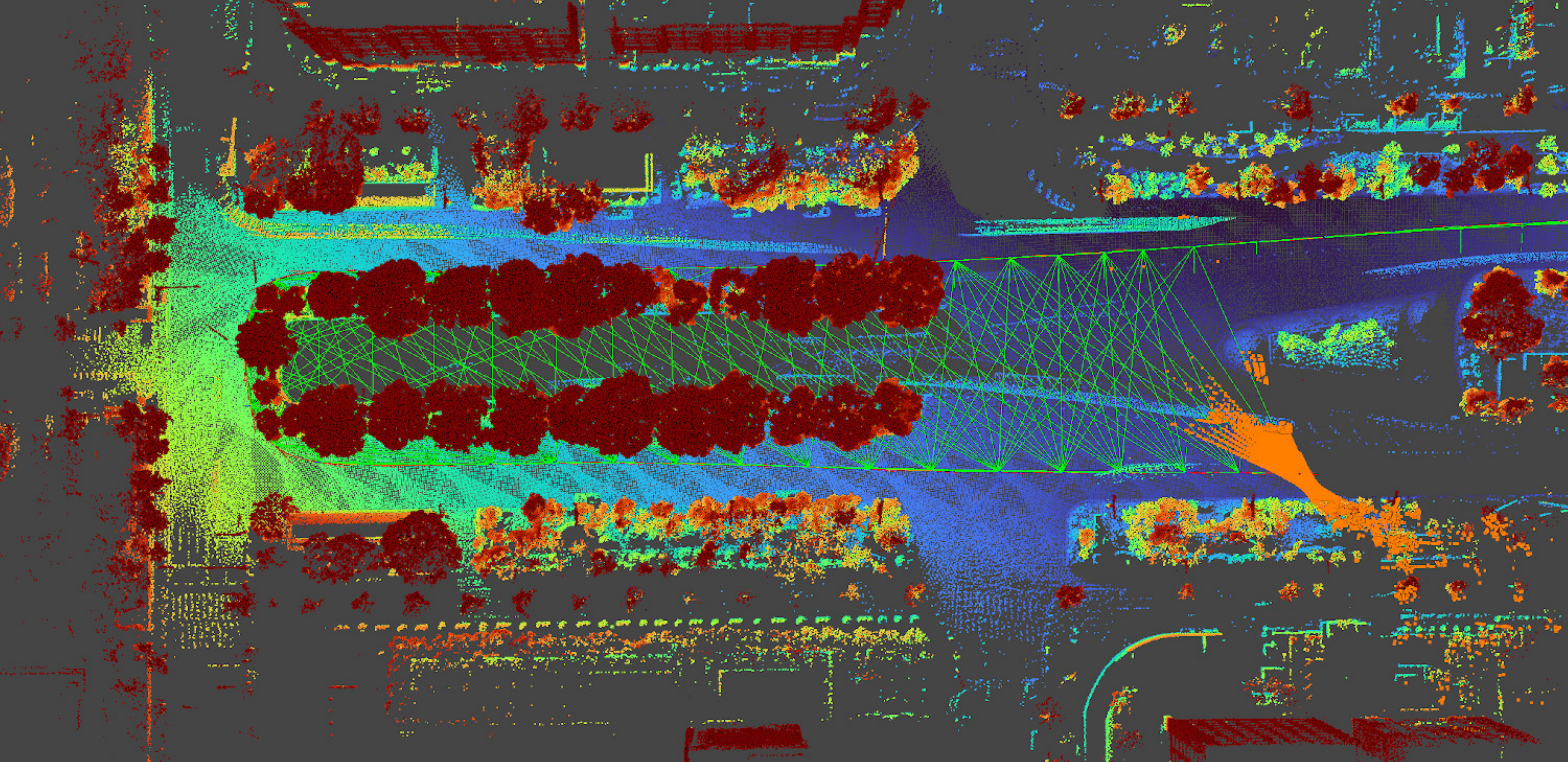}
  \subcaption{Constraints between small overlapping frames}
  \end{minipage}
  \begin{minipage}[b]{0.44\linewidth}
  \centering
  \includegraphics[width=\linewidth]{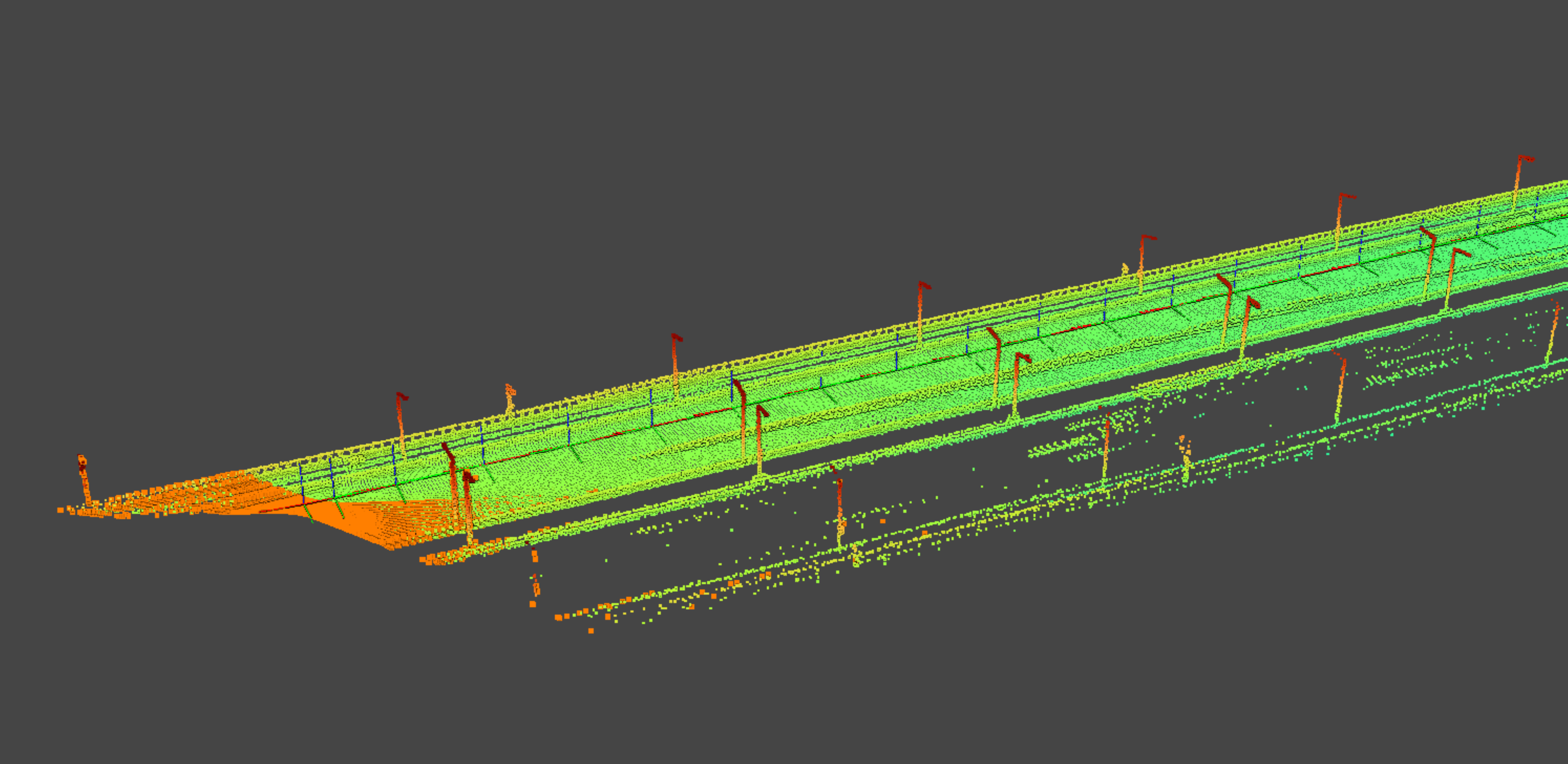}
  \subcaption{Feature-less highway environment}
  \end{minipage}
  \caption{Snapshots for the mapping process through the KAIST17 sequence. The orange points indicate the latest LiDAR scans.}
  \label{fig:kaist_17}
\end{figure*}

\begin{table}[tb]
  \centering
  \caption{Processing time through the KAIST07 sequence}
  \label{tab:proctime_kaist}
  \begin{tabular}{llc}
  \toprule
  Module & Process & Time [msec] \\
  \midrule
  \midrule
  \multirow{3}{*}{Preprocess}
                      & Downsampling       & 4.4 $\pm$ 1.0  \\
                      & kNN search         & 18.9 $\pm$ 3.1 \\
                      & Total              & 23.3 $\pm$ 4.0 \\
  \midrule
  \multirow{4}{*}{Odometry estimation}
                      & Deskew \& Cov.     & 6.7 $\pm$ 6.2   \\
                      & Optimization       & 21.0 $\pm$ 11.7 \\
                      & Keyframe update    & 13.5 $\pm$ 12.9 \\
                      & Total              & 41.3 $\pm$ 18.3 \\
  \midrule
  \multirow{5}{*}{Local mapping}
                      & Deskew \& Cov.     & 8.0 $\pm$ 7.6    \\
                      & Total (per-frame)  & 8.9 $\pm$ 8.0    \\ \cmidrule{2-3}
                      & Optimization       & 117.0 $\pm$ 48.8 \\
                      & Merging frames     & 5.0 $\pm$ 5.2    \\
                      & Total (per-submap) & 122.0 $\pm$ 49.7 \\
  \midrule
  \multirow{3}{*}{Global mapping}
                      & Factor creation    & 22.2 $\pm$ 21.1   \\
                      & Optimization       & 208.9 $\pm$ 120.0 \\
                      & Total              & 242.5 $\pm$ 136.1 \\
  \bottomrule
  \end{tabular}
\end{table}

\begin{figure}[tb]
  \centering
  \includegraphics[width=0.95\linewidth]{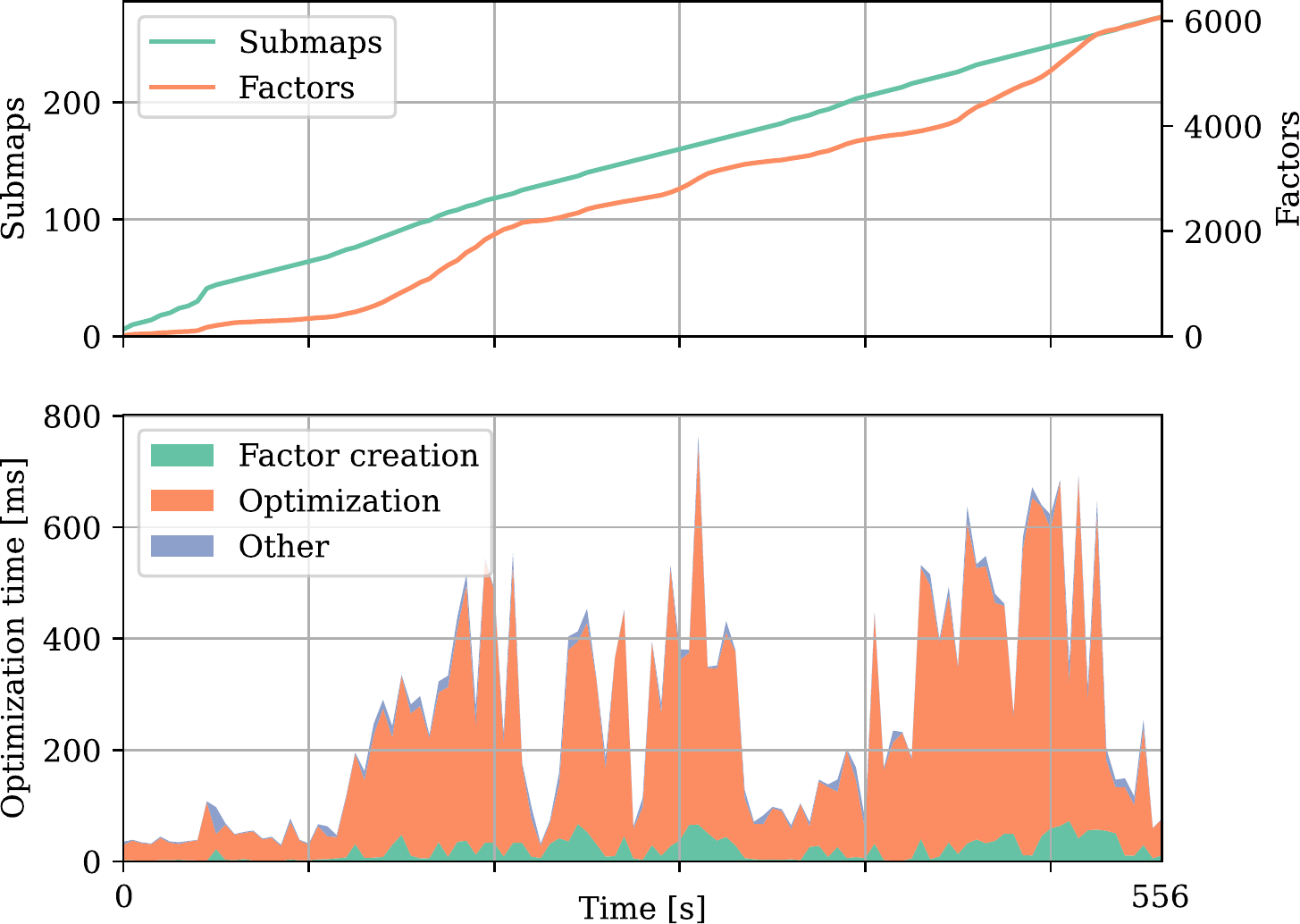}
  \caption{Number of submaps and matching cost factors and the processing time of the global optimization through the KAIST07 sequence.}
  \label{fig:proctime}
\end{figure}

To demonstrate that the proposed framework can robustly estimate the sensor trajectory in challenging situations, we conducted experiments on KAIST urban dataset \cite{Jeong2018}. In this dataset, a LiDAR (Velodyne VLP-16) was vertically mounted on the vehicle (see Fig. \ref{fig:kaist_urban}), and thus consecutive LiDAR point clouds only have a small overlap when the vehicle is running. It is often difficult to obtain sufficient geometrical constraints from LiDAR point clouds, and tight coupling of LiDAR and IMU constraints is inevitable.

Fig. \ref{fig:kaist_07} shows a mapping result for the KAIST07 sequence. We can see that the proposed method was able to create a consistent environmental map in the challenging setting thanks to the tightly coupled LiDAR-IMU fusion scheme. We can also see that the proposed framework aggressively creates matching cost factors between small overlapping frames, and an extremely dense factor graph is constructed.

Fig. \ref{fig:kaist_17} shows snapshots of a trial with the KAIST17 sequence. The proposed backend algorithm is very powerful and enables the creation of constraints between small overlapping frames, which  helps correcting trajectory estimation drift, as shown in Fig. \ref{fig:kaist_17} (a). It can also robustly estimate the sensor trajectory in a feature-less highway environment, as shown in Fig. \ref{fig:kaist_17} (b) \footnotemark[2]. Note that, without the IMU constraints, the global optimization corrupted on the highway environment due to insufficient geometrical features.

Through the KAIST07 sequence, the proposed framework ran approximately twice as fast as the real-time elapsed (20 FPS). Table \ref{tab:proctime_kaist} summarizes the processing times of each module in the proposed framework. The preprocessing and odometry estimation modules respectively took 23.3 ms and 41.3 ms per frame and were sufficiently faster than the real-time requirement (100 ms per frame). The submap optimization, which was performed approximately every 2 s, took 122.0 ms on average. The global map optimization, which was performed approximately every 5 s, took 242.5 ms on average. 

Fig. \ref{fig:proctime} shows how the global optimization time grew as the number of submaps and matching cost factors increased. While a massive amount of matching cost factors are created (over 6,000 factors), the global optimization converged in less than one second thanks to the incremental optimizer and GPU acceleration.

\section{Conclusions}

This paper presents a LiDAR-IMU mapping framework. The proposed framework comprises odometry estimation, local mapping, and global mapping modules, which are all based on the LiDAR-IMU tight coupling. For odometry estimation, an efficient keyframe mechanism and fixed-lag smoothing technique are used to achieve a low-drift estimation with a bounded computation cost. A new factor graph structure for the backend was proposed to realize tightly coupled LiDAR-IMU fusion. We validated the efficiency and accuracy of the proposed framework using the Newer College dataset and KAIST urban dataset.

\balance

\bibliographystyle{IEEEtran}
\bibliography{icra2022}

\begin{thebibliography}{10}
\providecommand{\url}[1]{#1}
\csname url@rmstyle\endcsname
\providecommand{\newblock}{\relax}
\providecommand{\bibinfo}[2]{#2}
\providecommand\BIBentrySTDinterwordspacing{\spaceskip=0pt\relax}
\providecommand\BIBentryALTinterwordstretchfactor{4}
\providecommand\BIBentryALTinterwordspacing{\spaceskip=\fontdimen2\font plus
\BIBentryALTinterwordstretchfactor\fontdimen3\font minus
  \fontdimen4\font\relax}
\providecommand\BIBforeignlanguage[2]{{%
\expandafter\ifx\csname l@#1\endcsname\relax
\typeout{** WARNING: IEEEtran.bst: No hyphenation pattern has been}%
\typeout{** loaded for the language `#1'. Using the pattern for}%
\typeout{** the default language instead.}%
\else
\language=\csname l@#1\endcsname
\fi
#2}}

\bibitem{Ye2019}
H.~Ye, Y.~Chen, and M.~Liu, ``Tightly coupled {3D} lidar inertial odometry and
  mapping,'' in \emph{{IEEE} International Conference on Robotics and
  Automation}.\hskip 1em plus 0.5em minus 0.4em\relax {IEEE}, May 2019, pp.
  3144--3150.

\bibitem{Qin2020}
C.~Qin, H.~Ye, C.~E. Pranata, J.~Han, S.~Zhang, and M.~Liu, ``{LINS}: A
  lidar-inertial state estimator for robust and efficient navigation,'' in
  \emph{{IEEE} International Conference on Robotics and Automation}.\hskip 1em
  plus 0.5em minus 0.4em\relax {IEEE}, May 2020, pp. 8899--8906.

\bibitem{Qin2018}
T.~Qin, P.~Li, and S.~Shen, ``{VINS}-mono: A robust and versatile monocular
  visual-inertial state estimator,'' \emph{{IEEE} Transactions on Robotics},
  vol.~34, no.~4, pp. 1004--1020, Aug. 2018.

\bibitem{koide_ral2021}
K.~Koide, M.~Yokozuka, S.~Oishi, and A.~Banno, ``Globally consistent {3D}
  {LiDAR} mapping with {GPU}-accelerated {GICP} matching cost factors,''
  \emph{{IEEE} Robotics and Automation Letters}, pp. 1--8, 2021.

\bibitem{Ramezani2020}
M.~Ramezani, Y.~Wang, M.~Camurri, D.~Wisth, M.~Mattamala, and M.~Fallon, ``The
  newer college dataset: Handheld {LiDAR}, inertial and vision with ground
  truth,'' in \emph{{IEEE}/{RSJ} International Conference on Intelligent Robots
  and Systems}.\hskip 1em plus 0.5em minus 0.4em\relax {IEEE}, Oct. 2020, pp.
  4353--4360.

\bibitem{Jeong2018}
J.~Jeong, Y.~Cho, Y.-S. Shin, H.~Roh, and A.~Kim, ``Complex urban {LiDAR} data
  set,'' in \emph{{IEEE} International Conference on Robotics and
  Automation}.\hskip 1em plus 0.5em minus 0.4em\relax {IEEE}, May 2018, pp.
  6344--6351.

\bibitem{Engel2018}
J.~Engel, V.~Koltun, and D.~Cremers, ``Direct sparse odometry,'' \emph{{IEEE}
  Transactions on Pattern Analysis and Machine Intelligence}, vol.~40, no.~3,
  pp. 611--625, Mar. 2018.

\bibitem{Stumberg2018}
L.~V. Stumberg, V.~Usenko, and D.~Cremers, ``Direct sparse visual-inertial
  odometry using dynamic marginalization,'' in \emph{{IEEE} International
  Conference on Robotics and Automation}.\hskip 1em plus 0.5em minus
  0.4em\relax {IEEE}, May 2018, pp. 2510--2517.

\bibitem{Campos2021}
C.~Campos, R.~Elvira, J.~J.~G. Rodriguez, J.~M.~M. Montiel, and J.~D. Tardos,
  ``{ORB}-{SLAM}3: An accurate open-source library for visual,
  visual{\textendash}inertial, and multimap {SLAM},'' \emph{{IEEE} Transactions
  on Robotics}, pp. 1--17, 2021.

\bibitem{liosam2020shan}
T.~Shan, B.~Englot, D.~Meyers, W.~Wang, C.~Ratti, and R.~Daniela, ``{LIO-SAM}:
  Tightly-coupled lidar inertial odometry via smoothing and mapping,'' in
  \emph{{IEEE}/{RSJ} International Conference on Intelligent Robots and
  Systems}.\hskip 1em plus 0.5em minus 0.4em\relax {IEEE}, Oct. 2020, pp.
  5135--5142.

\bibitem{Weiss2011}
S.~Weiss and R.~Siegwart, ``Real-time metric state estimation for modular
  vision-inertial systems,'' in \emph{{IEEE} International Conference on
  Robotics and Automation}.\hskip 1em plus 0.5em minus 0.4em\relax {IEEE}, May
  2011, pp. 4531--4537.

\bibitem{Indelman2013}
V.~Indelman, S.~Williams, M.~Kaess, and F.~Dellaert, ``Information fusion in
  navigation systems via factor graph based incremental smoothing,''
  \emph{Robotics and Autonomous Systems}, vol.~61, no.~8, pp. 721--738, Aug.
  2013.

\bibitem{Xu2021}
W.~Xu and F.~Zhang, ``{FAST}-{LIO}: A fast, robust {LiDAR}-inertial odometry
  package by tightly-coupled iterated kalman filter,'' \emph{{IEEE} Robotics
  and Automation Letters}, vol.~6, no.~2, pp. 3317--3324, Apr. 2021.

\bibitem{Li2021}
K.~Li, M.~Li, and U.~D. Hanebeck, ``Towards high-performance
  solid-state-{LiDAR}-inertial odometry and mapping,'' \emph{{IEEE} Robotics
  and Automation Letters}, vol.~6, no.~3, pp. 5167--5174, July 2021.

\bibitem{Palieri2021}
M.~Palieri, B.~Morrell, A.~Thakur, K.~Ebadi, J.~Nash, A.~Chatterjee,
  C.~Kanellakis, L.~Carlone, C.~Guaragnella, and A.~akbar Agha-mohammadi,
  ``{LOCUS}: A multi-sensor lidar-centric solution for high-precision odometry
  and 3d mapping in real-time,'' \emph{{IEEE} Robotics and Automation Letters},
  vol.~6, no.~2, pp. 421--428, Apr. 2021.

\bibitem{zhao2021super}
S.~Zhao, H.~Zhang, P.~Wang, L.~Nogueira, and S.~Scherer, ``{Super Odometry}:
  {IMU}-centric {LiDAR-Visual-Inertial} estimator for challenging
  environments,'' arXiv:2104.14938, 2021.

\bibitem{Shan2018}
T.~Shan and B.~Englot, ``{LeGO}-{LOAM}: Lightweight and ground-optimized lidar
  odometry and mapping on variable terrain,'' in \emph{{IEEE}/{RSJ}
  International Conference on Intelligent Robots and Systems}.\hskip 1em plus
  0.5em minus 0.4em\relax {IEEE}, Oct. 2018, pp. 4758--4765.

\bibitem{Reijgwart2020}
V.~Reijgwart, A.~Millane, H.~Oleynikova, R.~Siegwart, C.~Cadena, and J.~Nieto,
  ``Voxgraph: Globally consistent, volumetric mapping using signed distance
  function submaps,'' \emph{{IEEE} Robotics and Automation Letters}, vol.~5,
  no.~1, pp. 227--234, Jan. 2020.

\bibitem{Oleynikova2017}
H.~Oleynikova, Z.~Taylor, M.~Fehr, R.~Siegwart, and J.~Nieto, ``Voxblox:
  Incremental 3d euclidean signed distance fields for on-board {MAV}
  planning,'' in \emph{{IEEE}/{RSJ} International Conference on Intelligent
  Robots and Systems}.\hskip 1em plus 0.5em minus 0.4em\relax {IEEE}, Sept.
  2017, pp. 1366--1373.

\bibitem{vgicp}
K.~Koide, M.~Yokozuka, S.~Oishi, and A.~Banno, ``Voxelized {GICP} for fast and
  accurate {3D} point cloud registration,'' in \emph{{IEEE} International
  Conference on Robotics and Automation}.\hskip 1em plus 0.5em minus
  0.4em\relax {IEEE}, May 2021.

\bibitem{Segal2009}
A.~Segal, D.~Haehnel, and S.~Thrun, ``Generalized-{ICP},'' in \emph{Robotics:
  Science and Systems {V}}.\hskip 1em plus 0.5em minus 0.4em\relax Robotics:
  Science and Systems Foundation, June 2009, pp. 435--443.

\bibitem{Forster2017}
C.~Forster, L.~Carlone, F.~Dellaert, and D.~Scaramuzza, ``On-manifold
  preintegration for real-time visual--inertial odometry,'' \emph{{IEEE}
  Transactions on Robotics}, vol.~33, no.~1, pp. 1--21, Feb. 2017.

\bibitem{Levenberg1944}
K.~Levenberg, ``A method for the solution of certain non-linear problems in
  least squares,'' \emph{Quarterly of Applied Mathematics}, vol.~2, no.~2, pp.
  164--168, July 1944.

\bibitem{MurArtal2017a}
R.~Mur-Artal and J.~D. Tardos, ``{Visual-Inertial} monocular {SLAM} with map
  reuse,'' \emph{{IEEE} Robotics and Automation Letters}, vol.~2, no.~2, pp.
  796--803, Apr. 2017.

\bibitem{Kaess2011}
M.~Kaess, H.~Johannsson, R.~Roberts, V.~Ila, J.~J. Leonard, and F.~Dellaert,
  ``{iSAM}2: Incremental smoothing and mapping using the bayes tree,''
  \emph{International Journal of Robotics Research}, vol.~31, no.~2, pp.
  216--235, Dec. 2011.

\bibitem{Zhang2018}
Z.~Zhang and D.~Scaramuzza, ``A tutorial on quantitative trajectory evaluation
  for visual(-inertial) odometry,'' in \emph{{IEEE}/{RSJ} International
  Conference on Intelligent Robots and Systems}.\hskip 1em plus 0.5em minus
  0.4em\relax {IEEE}, Oct. 2018, pp. 7244--7251.

\end{thebibliography}

\end{document}